\documentclass[runningheads]{llncs}
\usepackage[T1]{fontenc}
\usepackage{cite}
\usepackage{amsmath,amssymb,amsfonts}
\usepackage{algorithm}
\usepackage{algpseudocode}
\usepackage{siunitx} 
\usepackage{graphicx}
\usepackage{textcomp}
\usepackage{xcolor}
\usepackage{color}
\usepackage[table]{xcolor}
\usepackage{tcolorbox}
\usepackage{booktabs}
\usepackage{multirow}
\graphicspath{{./images/}} 
\usepackage{wrapfig}       
\usepackage{caption}       
\usepackage{subcaption}    
\usepackage{listings}
\usepackage{paralist} 
\usepackage[hidelinks]{hyperref} 

\usepackage[all]{hypcap}
\usepackage[nameinlink,capitalise]{cleveref} 
\def\BibTeX{{\rm B\kern-.05em{\sc i\kern-.025em b}\kern-.08em
    T\kern-.1667em\lower.7ex\hbox{E}\kern-.125emX}}

\makeatletter
\renewcommand\section{\@startsection{section}{1}{\z@}%
                       {-12\p@ \@plus -4\p@ \@minus -4\p@}%
                       {6\p@ \@plus 4\p@ \@minus 4\p@}%
                       {\normalfont\Large\bfseries\boldmath
                        \rightskip=\z@ \@plus 8em\pretolerance=10000 }}
\renewcommand\subsection{\@startsection{subsection}{2}{\z@}%
                       {-10\p@ \@plus -4\p@ \@minus -4\p@}%
                       {4\p@ \@plus 4\p@ \@minus 4\p@}%
                       {\normalfont\large\bfseries\boldmath
                        \rightskip=\z@ \@plus 8em\pretolerance=10000 }}
\makeatother

\begin{document}

\title{The Pursuit of Diversity: Multi-Objective Testing of Deep Reinforcement Learning Agents
}
\titlerunning{The Pursuit of Diversity}
\author{Antony Bartlett\orcidID{0009-0008-2654-8556}, Cynthia Liem\orcidID{0000-0002-5385-7695}, Annibale Panichella\orcidID{0000-0002-7395-3588}
}
\institute{Delft University of Technology, Delft, The Netherlands \\
 \email{a.j.bartlett@tudelft.nl}, \email{c.c.s.liem@tudelft.nl}, \email{a.panichella@tudelft.nl} 
}

\maketitle

\newcommand{\eg}{\textit{e.g.,}}
\newcommand{\Eg}{\textit{E.g.,}}
\newcommand{\ie}{\textit{i.e.,}}
\newcommand{\Ie}{\textit{I.e.,}}
\newcommand{\etal}{\textit{et al.}}
\newcommand{\etc}{\textit{etc.}}
\newcommand{\wrt}{\textit{w.r.t.}}
\newcommand{\cfr}{\textit{cfr.}}
\newcommand{\Cfr}{\textit{Cfr.}}
\newcommand{\viz}{\textit{viz.}}
\newcommand{\aka}{\textit{a.k.a.}}
\newcommand{\cf}{\textit{cf.}}
\newcommand{\Cf}{\textit{Cf.}}
\newcommand{\indago}{\texttt{INDAGO}}
\newcommand{\nexus}{\texttt{INDAGO-Nexus}}
\newcommand{\nexusga}{\nexus{}-\texttt{GA}}
\newcommand{\nsgatwo}{\texttt{NSGA-II}}
\newcommand{\agemoea}{\texttt{AGE-MOEA}}
\newcommand{\atwelve}{\mathrm{\hat{A}}_{12}}
\newcommand{\pn}{%
  \bfseries%
  \ifnum\value{ALG@line} < 10 \hphantom{0}%
  \else\relax\fi%
  \theALG@line%
}

\definecolor{lightgrey}{rgb}{0.2,0.2,0.2}
\definecolor{darkgrey}{rgb}{0.0,0.0,0.0}

\newtheorem{mydef}{Definition}
\newtheorem{probdef}{Problem Definition}

\captionsetup{
  justification = centering
}

\vspace{-1em}
\begin{abstract}\label{sec:abstract}

Testing deep reinforcement learning (DRL) agents in safety-critical domains requires discovering diverse failure scenarios. Existing tools such as \indago{} rely on single-objective optimization focused solely on maximizing failure counts, but this does not ensure discovered scenarios are diverse or reveal distinct error types. We introduce \nexus{}, a multi-objective search approach that jointly optimizes for failure likelihood and test scenario diversity using multi-objective evolutionary algorithms with multiple diversity metrics and Pareto front selection strategies. We evaluated \nexus{} on three DRL agents: humanoid walker, self-driving car, and parking agent. On average, \nexus{} discovers up to 83\% and 40\% more unique failures (test effectiveness) than \indago{} in the SDC and Parking scenarios, respectively, while reducing time-to-failure by up to 67\% across all agents.

\end{abstract}

\section{Introduction}\label{sec:intro}

Reinforcement Learning (RL)\cite{Sutton1998}, and more recently Deep Reinforcement Learning (DRL)\cite{mnih2013playingatarideepreinforcement}, have emerged as powerful tools for automating complex tasks in dynamic and uncertain environments~\cite{simonyan2014deepinsideconvolutionalnetworks, MnihDRL2015}.
DRL agents learn through trial-and-error interactions with their environment and are increasingly deployed in Cyber-Physical Systems (CPS), such as self-driving cars (SDCs) and humanoid robots~\cite{bloom2023decentralizedmultiagentreinforcementlearning}, where safety, reliability, and adaptability are critical.

Ensuring the reliability of DRL-based systems remains a significant challenge~\cite{uesato2018rigorousagentevaluationadversarial, MatteoIndago2024}. Unlike conventional software, DRL agents are opaque, stochastic, and highly sensitive to reward signals. Their behavior depends not only on a learned policy but also on vast environment and action spaces. As a result, edge-case scenarios—those most likely to trigger failures—are difficult to identify, and traditional testing approaches often fall short~\cite{uesato2018rigorousagentevaluationadversarial}. Real-world incidents involving autonomous agents~\cite{theguardiantesla2024, newatlus2025humanoid, electrek2025tesla} underscore the consequences of failing to detect such rare conditions during testing.

Simulation-based testing is a widely adopted technique for evaluating DRL agents, especially in safety-critical domains such as autonomous agents. It allows developers to generate and execute test scenarios in a controlled and reproducible setting, substantially reducing the time, cost, and safety risks of real-world testing~\cite{kaur2021surveysimulatorstestingselfdriving, giubilato2020simulationframeworkmobilerobots}.
A test scenario refers to a specific configuration of the environment—including initial agent state, obstacle placement, and task-specific parameters—that defines the context in which the agent is evaluated.
Despite its advantages, simulation-based testing still remains resource-intensive, and exhaustive exploration of the test input space is often infeasible due to its combinatorial complexity.

To address these challenges, Biagiola and Tonella~\cite{MatteoIndago2024} proposed the use of surrogate models to guide test generation for DRL agents. Their framework, \indago{}, builds a surrogate model from training logs to predict whether a given test scenario is likely to cause agent failure—without executing it in simulation. \indago{} combines this model with two search strategies: Hill Climbing (HC) and a single-objective Genetic Algorithm (GA).
Their empirical evaluation across multiple DRL tasks showed that (1) surrogate models substantially reduced simulation time, and (2) both HC and GA outperformed earlier random-search-based methods~\cite{uesato2018rigorousagentevaluationadversarial} in identifying more failure-inducing scenarios.

While \indago{} effectively increases the number of failing scenarios, it focuses solely on failure likelihood, overlooking diversity. This is a key limitation—generating more failures does not guarantee that they are distinct. Diversity is crucial in testing as it promotes broader coverage of the agent’s operational space and improves the chances of uncovering unique faults, helping to address the \textit{curse of rarity} problem~\cite{liu2022curserarityautonomousvehicles}.

In this paper, we propose \nexus{}, a multi-objective search framework for DRL testing that jointly optimizes for failure probability and scenario diversity. We investigate:
\begin{enumerate}\label{sec:rq1}
    \item[\textbf{RQ1}:] \textit{How does multi-objective search perform in finding diverse failures?}
    \item[\textbf{RQ2}:] \textit{How effective is \nexus{} compared to state-of-the-art \indago{}?}
\end{enumerate}

To tackle this inherently multi-objective problem, we employed multi-objective evolutionary algorithms (MOEAs). MOEAs are well-suited for this task as they efficiently explore trade-offs and produce a set of Pareto-optimal solutions~\cite{deb2011multi, haq2022efficient,ishibuchi2008evolutionary}.
We experimented with two diversity metrics (Euclidean distance and PCA-based clustering), two MOEAs (NSGA-II~\cite{deb2002fast} and AGE-MOEA~\cite{panichella2019agemoea}), and two Pareto-front selection strategies (highest failure likelihood and knee point~\cite{zhang2014knee}). We benchmarked \nexus{} on three DRL agents: self-driving car, humanoid walker, and parking agent.

Our results indicate \nexus{} discovers up to 83\% and 40\% more unique failures than \indago{} in the SDC and Parking scenarios, respectively, while reducing time-to-failure by up to 67\%. Knee-point selection consistently produces the best results across most configurations, while different diversity metrics show varying effectiveness depending on the scenario.

Therefore, our contributions are:
\begin{itemize}
    \item We propose \nexus{}, a multi-objective approach to generate diverse test environments for DRL agents.
    \item We demonstrate \nexus{} discovers up to 83\% more unique failures than \indago{} while reducing time-to-failure by up to 67\%.
    \item We provide a complete replication package with implementation and experimental data\footnote{\url{10.6084/m9.figshare.29204687}}.
\end{itemize}

\section{Background and Related Work}
\label{sec:background}

This section describes the key background concepts used in this work and summarizes the related work.

\subsection{Deep Reinforcement Learning}
\label{subsec:drl}
Deep Reinforcement Learning (DRL) extends Reinforcement Learning (RL) by using deep neural networks as function approximators to handle high-dimensional inputs (e.g., images) and continuous action spaces (e.g., steering angles in autonomous driving)~\cite{MnihDRL2015}. Instead of explicitly representing policies or value functions, DRL learns them directly from raw observations and rewards using deep learning architectures.

This capability makes DRL particularly suitable for complex tasks in cyber-physical systems such as self-driving cars (SDCs), where the environment is dynamic and the decision space is large. In this work, we focus on \textit{model-free} DRL algorithms, where agents learn directly from interactions without an explicit model of the environment. 
For example, the DRL agent for the parking scenario we consider in our case study (see Section~\ref{sec:evaluation}) was trained with \textit{Proximal Policy Optimization (PPO)}~\cite{ppo2017schulman}, a widely adopted policy-gradient method known for its stability and sample efficiency.

\begin{quote}
\textit{The agent’s goal is to maximize the cumulative reward it receives in the long run}~\cite{Sutton1998}.
\end{quote}

\subsection{Black-box Testing}\label{subsec:diversity}
In black-box testing, diversity is a common strategy for exposing faults by varying inputs and observing outputs~\cite{guidotti2018survey}. For DRL agents, we focus on input diversity—variations in environment configurations—to guide test generation without costly simulations, combined with a surrogate to estimate failure likelihood.

Related work on diversity-based testing for deep neural networks includes approaches that optimize for input diversity to improve fault detection. Aghababaeyan et al.~\cite{aghababaeyan2021blackbox} proposed black-box testing techniques that use test case diversity to generate diverse test inputs for deep neural networks. While their work demonstrates the effectiveness of diversity-based approaches, their diversity metrics are primarily designed for image inputs rather than the tabular data that characterizes our MLP-based surrogate models. Our work adapts the concept of diversity-driven testing to the specific context of DRL environment configurations, which are typically represented as structured tabular data with both continuous and categorical features.

\subsection{Surrogates}\label{subsec:surrogate}

\begin{figure}[t]
	\centering
	\begin{subfigure}{0.48\columnwidth}
		\centering
		\begin{tcolorbox}[colback=gray!5!white, colframe=gray!75!black, width=0.95\textwidth, arc=2mm, boxrule=0.5pt]
		\scriptsize
		\begin{verbatim}
"env_config": {
    "goal_lane_idx": 0,
    "heading_ego": 0.96,
    "parked_vehicles_lane_indices": [
        1, 3, 6, 8, 9,
        10, 11, 12, 14, 18
    ],
    "position_ego": [
        1.83, -4.96
    ]
}
		\end{verbatim}
		\end{tcolorbox}
		\caption{Environment configuration}
		\label{fig:environment}
	\end{subfigure}%
	\hfill
	\begin{subfigure}{0.48\columnwidth}
		\centering
		\includegraphics[width=0.9\linewidth]{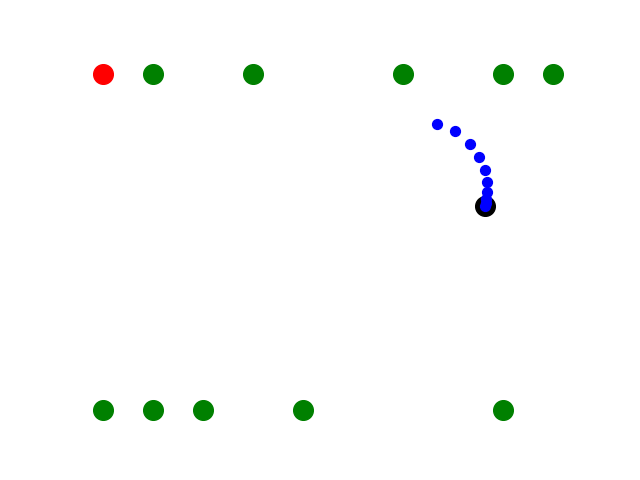}
		\caption{Agent behavior visualization}
		\label{fig:output}
	\end{subfigure}%
	\caption{Parking scenario showing (a) environment configuration details used by DRL as a starting point, and (b) the DRL agent behavior. In (b), the red dot indicates the target parking spot, green dots represent parked cars, the black dot marks the agent's starting position, and the blue dots trace its trajectory.}
	\label{fig:parking_scenario}
\end{figure}

Surrogate models are lightweight machine learning approximations that learn to predict the behavior or outcomes of computationally expensive processes based on historical data. In the context of DRL testing, these models have proven particularly effective at reducing evaluation costs~\cite{MatteoIndago2024}. Specifically, given an environment configuration (Figure~\ref{fig:environment}), a surrogate model can predict the failure probability in the range $[0,1]$, thereby enabling efficient test generation without requiring costly simulation executions.

\subsection{Testing DRL Agents}
\label{subsec:existing}

The most directly related work is by Biagiola and Tonella~\cite{MatteoIndago2024}, who introduced \indago{}, a test generation framework that leverages surrogate models trained on DRL logs to predict whether a given test environment is likely to cause failure. These surrogates return a continuous score in $[0,1]$, representing the estimated failure probability of an environment. \indago{} combines this prediction with search-based strategies, namely Hill Climbing~\cite{hc2006selman} and Genetic Algorithms~\cite{goldberg1989ga}, to efficiently explore the input space.

Our work builds on \indago{} by introducing a second optimization objective: input diversity. This extension aims to increase the diversity of the discovered failures, thereby improving coverage of the agent’s operational space.

\section{Approach}\label{sec:approach}

The goal of our approach is to \textit{generate test scenarios that are both failure-inducing and diverse}, increasing the likelihood of uncovering unique faults in DRL agents. 
We build upon the \indago{} framework~\cite{MatteoIndago2024}, which uses a surrogate model trained on DRL training logs to predict whether a given environment is likely to cause a failure. While \indago{} focuses solely on failure likelihood, we extend it to a \textit{multi-objective test generation framework} that also accounts for scenario diversity—crucial for uncovering unique faults.
Hence, we formulate the problem of generative diverse failure-inducing test scenarios as follows:

\begin{probdef}\label{prob:environment}
\textit{
 Given a DRL agent $A$,  a set of test scenarios $E$, and a surrogate model $s: E \rightarrow [0,1]$ that estimates the likelihood of failure for a given $e$, the goal is to generate a set of test scenarios $E_{\text{sel}} \subseteq E$ that maximize the likelihood of $A$ to fail while maximizing diversity among the selected scenarios. More formally, the goal is to find a set of environments $E_{\text{sel}}$ that optimizes the following objectives:
}
\begin{equation}
\label{eq:mo-objective}
\left\{
\begin{aligned}
\max O_1 &= s(e) \quad &\text{(Failure Likelihood)} \\
\max O_2 &=  Div(e, E_{\text{sel}}) \quad &\text{(Input Diversity)} \\
\end{aligned}
\right.
\end{equation}
\end{probdef}

\noindent where $O_1$ (Failure Likelihood) is the predicted likelihood of failure for a given environment $e$; and $O_2$ (Input Diversity) is a measure of how diverse the selected environments are compared to the other test scenarios in $E_{\text{sel}}$.

In the following, we describe the surrogate model, diversity metrics, and the multi-objective optimization framework.

\subsection{Surrogate models}
\label{subsec:our_surrogate}

Following the prior work by Biagiola and Tonella~\cite{MatteoIndago2024}, we adopt a lightweight surrogate model to estimate the failure likelihood of a test scenario without executing it. Specifically, we utilize a multi-layer perceptron (MLP) classifier trained on data collected during the DRL agent’s training episodes. The MLP architecture is well-suited for this task due to the relatively small training dataset (up to 10k labeled environments) and the low-dimensional nature of environment feature vectors (e.g., 24-dimensional in the parking scenario).
The surrogate outputs $s(e) \in [0,1]$, representing failure probability for environment $e$.

\subsection{Diversity Metrics}\label{subsec:divmetrics}

We explore two diversity methods: \textbf{Euclidean distance} computes the mean distance between a new scenario and previously selected ones, and \textbf{PCA-based clustering}~\cite{ding2004pca} measures distance to the nearest cluster centroid in reduced dimensionality space. Both methods handle variable-length features through one-hot encoding and zero-padding. 

\subsection{Multi-Objective Evolutionary Testing}\label{subsec:nexus}
To generate a diverse set of failure-inducing test scenarios, \nexus{} employs MOEAs that evolve environment configurations to simultaneously maximize (1) the likelihood of failure, as predicted by a surrogate model, and (2) diversity with respect to a growing archive of previously selected tests. We use two MOEAs: NSGA-II~\cite{deb2002fast} and AGE-MOEA~\cite{panichella2019agemoea} to optimize both failure likelihood and diversity.

Algorithm~\ref{alg:evolution} outlines this process. The search begins with a population of size PS. Each individual is evaluated without test execution using heuristics described in Sections~\ref{subsec:our_surrogate} and~\ref{subsec:diversity}. The algorithm proceeds for $G$ generations. In each generation, two parents are selected (Lines~9-10), the crossover is applied with probability $CR$ (Line~11), and both offspring are mutated (Lines~14-15). Offspring are added to the new population (Line~16), and elitism selects the best individuals for the next generation (Line~18).

\begin{algorithm}[t]
\caption{Multi-Objective Test Generation for DRL Agents}
\label{alg:evolution}
\scriptsize
\begin{algorithmic}[1]
\Require $s$, classifier (surrogate model);
\Require TR, test runs;
\Require G, generations;
\Require PS, population size;
\Require CR, crossover rate;
\Require $E_{train}$, set of environment configurations where the DRL agent failed during training;
\Ensure $E_{sel}$, the archive storing a diverse set of likely-failing test scenarios.
\State currentTestRun $\gets 0$;
\While {$currentTestRun \leq TR$}
\State population $\gets$ GENERATE-POPULATION(PS, $E_{train}$);
\State COMPUTE-OBJECTIVES(population, $s$, $E_{sel}$);

\State currentGeneration $\gets 0$;
\While {$currentGeneration \leq G$}
 \State newPop $\gets \emptyset$ ;
    
 \While {$|$newPop$|$ $< PS$}
 \State $p_1 \gets$ SELECTION(population)
 \State $p_2 \gets$ SELECTION(population)

 \If {getRandomFloat() $< CR$}
 \State $o_1, o_2 \gets$ crossover($p_1$, $p_2$)
 \EndIf
        
 \State $o_1 \gets$ MUTATE($o_1$)
 \State $o_2 \gets$ MUTATE($o_2$)

 \State newPop $\gets$ newPop $\cup \{o_1, o_2\}$
 \EndWhile
    
 \State population $\gets$ ELITISM($\gets$ newPop $\cup$ population);

 \If {stagnation detected $or$ $currentGeneration \geq G$}
 \State $E_{sel} \gets E_{sel} \cup$ GET-BEST-INDIVIDUAL(population) 
 
 \EndIf
\EndWhile
\EndWhile
\State \Return $E_{sel}$
\end{algorithmic}
\end{algorithm}

To avoid stagnation, \nexus{} monitors hypervolume changes across generations. If improvement falls below a threshold, or we have reached the maximum number of generations $G$, the best individual is archived (Line~20). To explore new regions of space, the process continues until the total test runs $TR$ is met. The resulting archive $E_{sel}$ contains diverse, likely-failing scenarios, which are then executed on the DRL agent for validation.

While \nexus{} builds on established MOEAs like NSGA-II~\cite{deb2002fast} and AGE-MOEA~\cite{panichella2019agemoea}, it introduces several domain-specific adaptations: failure-based population seeding, surrogate-based fitness estimation, diversity measured against an evolving archive, and a stagnation-aware reset strategy for enhanced exploration. We describe these customizations in the following paragraphs.

\textbf{Population Initialization}. 
We seed the initial population with environments where the DRL agent failed during training ($E_{train}$), biasing the search towards failure-prone regions~\cite{MatteoIndago2024}.

\textbf{Selection}.
Selection determines which individuals are chosen for reproduction. \nexus{} uses binary tournament selection based on Pareto dominance~\cite{deb2002fast}, and the non-dominated sorting in particular.

\textbf{Objective Calculation}.
Each individual is evaluated without test execution. The surrogate model $s$ estimates failure likelihood, while diversity is computed with respect to the archive $E_{sel}$ using either Euclidean distance or PCA-based clustering metrics as described in Section~\ref{subsec:divmetrics}.

\textbf{Crossover and Mutation}.
\nexus{} applies single-point crossover (Line 12) probabilistically based on the crossover rate $CR$, randomly selecting a crossover point in the feature vector and swapping segments between parents.

Mutation (Lines 14-15) is \textit{always} applied to both offspring and uses \textit{saliency-based feature selection}~\cite{simonyan2014deepinsideconvolutionalnetworks}, which computes the gradient of the surrogate model's output with respect to each input feature. The saliency determines which features impact the surrogate's prediction, guiding mutation decisions.
For \textit{fixed-length features} (position, heading), we apply polynomial mutation~\cite{deb2011multi} with type-specific handling (real-valued, integer, binary, categorical). For \textit{variable-length features} (parked vehicle lists), we randomly apply removal, addition, or modification operations. A repair operator ensures validity after mutations.

\textbf{Elitism}. 
We retain the best individuals using NSGA-II's fast non-dominated sorting with crowding distance~\cite{deb2002fast} or AGE-MOEA's adaptive survival score~\cite{panichella2019agemoea}. 

\textbf{Stagnation and Archive Update.} 
\nexus{} maintains an archive $E_{sel}$ of selected scenarios. Stagnation is detected by monitoring the change in hypervolume of the Pareto front $P_g$ at generation $g$ relative to a reference point $\mathbf{r}$ (Equation~\ref{eq:motermination}). The hypervolume metric reflects both diversity and convergence by measuring the volume dominated by $P_g$ with respect to $\mathbf{r}$:
\begin{equation}\label{eq:motermination}
\small
\left| HV(P_g, \mathbf{r}) - HV(P_{g - n_{\text{last}}}, \mathbf{r}) \right| < \text{tol}
\end{equation}

When stagnation is detected, \nexus{} selects a representative individual from the Pareto front using two strategies: \textit{Maximum Failure Likelihood} (highest predicted failure probability) or \textit{Knee Point} (best trade-off between failure likelihood and diversity)~\cite{zhang2014knee}.

\section{Empirical Evaluation}\label{sec:evaluation}

Our empirical evaluation aims to answer the following research questions:
\begin{enumerate}\label{sec:rq_eval}
    \item[\textbf{RQ1}:] \textit{How does multi-objective search perform in finding diverse failures?}
    \item[\textbf{RQ2}:] \textit{How effective is \nexus{} compared to state-of-the-art \indago{}?}
\end{enumerate}

\subsection{Case Studies}\label{subsec:casestudies}

We evaluate \nexus{} on three diverse DRL agents from Biagiola and Tonella~\cite{MatteoIndago2024}, each representing different application domains and complexity:

\textbf{Parking.} The first environment we consider is the \texttt{parking} environment~\cite{highway-env}. In this scenario, the DRL agent begins at a specific location with a defined heading direction. The agent must park in a designated goal lane while avoiding collisions with any of the already parked vehicles. An example of this environment is illustrated in Figure~\ref{fig:environment}. The environment consists of 24 elements, including goal lane, heading, 20 parking slots, and the $x$, $y$ coordinates of the starting position.

\textbf{Humanoid.} Next, we examine the \texttt{Humanoid} environment, one of the more challenging environments from the MuJoCo simulator~\cite{Todorov2012MuJoCoAP}. In this scenario, the agent must control a bipedal robot to walk on a smooth surface within a 3D space. The environment configuration of the Humanoid robot consists of two arrays: joint position and joint velocity.  The joint position array contains 24 elements representing the positions and rotations of the robot's joints, while the joint velocity array contains the linear and angular velocities of the joints.

\textbf{Self-Driving Car.} The final environment we consider is the Self-Driving Car (SDC), developed using the DonkeyCar simulator~\cite{Kramer2021}. In this environment, the DRL agent must navigate a car from the starting point of a track to the end without leaving track boundaries. The environment configuration consists of a list with 12 pairs, where each pair comprises a command and a value. For example, a command could be a left or right turn, with the value specifying the turn length.

\subsection{Baseline}\label{subsec:baseline}

To ensure fair comparison of all approaches, we utilize the original models from the \indago{} paper. For our baseline, we employed the GA from the original \indago{} tool. This approach uses the predictive value of the surrogate model as a fitness value to gradually mutate environments in the search for failures. We executed this with the \texttt{saliency\_failure} test policy, which ensures the initial population is created from previously failing environments sourced from the training data, incorporating saliency (as discussed in Section~\ref{sec:approach}) during crossover and mutation. We selected this method as it demonstrated the highest rate of failure in the original study~\cite{MatteoIndago2024} against other approaches (\texttt{random search} and \texttt{hill climbing}). Due to the stochastic nature of evolutionary search, we executed the baseline $50$ times, applying the process across all three agent scenarios.

\subsection{Implementation and Parameter Settings}\label{subsec:parameters}

We implemented \agemoea{} and \nsgatwo{} using Pymoo v0.6.1~\cite{pymoo}. Experiments used 50 generations, population size 50, crossover rate 0.75, with stagnation-based termination ($tol=5 \times 10^{-6}$, $n_{\text{last}}=10$). The reference point for hypervolume calculation is set to $r = [1.2, 20.2]$, representing the upper bounds for the objectives. Experiments were conducted within Docker containers on a machine equipped with an AMD EPYC 7713 64-Core CPU (2.6 GHz), 256 threads, and an NVIDIA A40 GPU (48GB GDDR6) for tensor-based computation.

\subsection{Evaluation Criteria}\label{subsec:extraction}

We measure: (1) \textit{unique failures} via PCA+K-means clustering of execution traces, (2) \textit{input/output diversity} using entropy across clusters, and (3) \textit{time-to-failure (TTF)} efficiency. Statistical significance is assessed using Wilcoxon rank-sum test ($\alpha = 0.05$) and Vargha–Delaney effect size.

\subsection{Failure Detection and Diversity Assessment}\label{subsec:evaluation_criteria}

To assess the effectiveness of our multi-objective search strategies (\textbf{RQ1}), we executed both \nsgatwo{} and \agemoea{} across all combinations of diversity metrics and Pareto front selection strategies.

We focused on three aspects: the number of unique failures discovered, the diversity of these failures, and the efficiency in detecting them. We define a failure as a test scenario in which the DRL agent violates its task specification (e.g., crashing, falling, or missing the goal). To measure \textit{unique failures}, we cluster the execution traces (i.e., output trajectories) of failing tests using Principal Component Analysis (PCA) followed by K-means. Each resulting cluster corresponds to a distinct failure type, providing a behavior-grounded notion of uniqueness. Since trajectory lengths vary, we normalize them via zero-padding. For the Parking and SDC scenarios, we track the agent’s position over time, while in the Humanoid scenario, we monitor the robot’s vertical movement to detect falls.

To determine the optimal number of clusters $K^*$, we apply silhouette analysis and increase $K$ only if the silhouette score improves by at least 20\%, reducing the effect of noisy improvements. This clustering method is used both to compute output diversity and to count unique failures. It is important to distinguish this post hoc clustering from the PCA-based metric used within the search itself (Section~\ref{subsec:divmetrics}). The former relies on test execution results while the latter (our objective) relies on the input features (i.e., without running the tests).

We also evaluate the diversity of failures from both input and output perspectives. \textit{Output diversity} is measured using the same clustering of output trajectories used to identify unique failures. For \textit{input diversity}, we cluster the failing environment configurations using PCA and K-means, providing insight into the structural variety of the input scenarios, regardless of the agent’s behavior. In both cases, we calculate two diversity metrics: \textit{unique failures}, which reflects how many clusters are populated, and \textit{entropy}, which quantifies how evenly failures are distributed across clusters. We follow the definitions provided in the \indago{} framework~\cite{MatteoIndago2024}.

To assess \textit{test efficiency}, we compute the average time required to find the first failing test case. This metric reflects how quickly each approach exposes failures. The same metrics are then used in the comparison between \nexus{} and the baseline \indago{} (\textbf{RQ2}). To assess statistical significance, we apply the Wilcoxon rank-sum test~\cite{conover1998practical} with a confidence level of $\alpha = 0.05$, and we report the Vargha–Delaney $\atwelve{}$ statistics for the effect size~\cite{vargha2000critique}.

\section{Results}\label{sec:results}

This section presents the results of our study, addressing in turn each research question introduced earlier.

\begin{table}[!ht]
\caption{Performance metrics across different configurations for all three agents, displaying median and Interquartile Range (IQR) across 50 runs. (Gray cells indicate top values for each agent and approach.)}
\label{tab:all_results}
\centering
\resizebox{.9\textwidth}{!}{
\begin{tabular}{lccccc}
\multicolumn{6}{c}{\textbf{(a) Parking Agent}} \rule{0pt}{3ex}\\
\toprule
 & \multicolumn{2}{c}{\textbf{Failures}} & \multicolumn{2}{c}{\textbf{Entropy}} & \\
\cmidrule(lr){2-3} \cmidrule(lr){4-5}
\textbf{Configuration} & \textbf{\#Total} & \textbf{\#Unique} & \textbf{Input} & \textbf{Output} & \textbf{TTF} \\
\midrule
\multicolumn{6}{l}{\textbf{AGE-MOEA}} \\
PCA/Knee & \cellcolor{lightgray} 10 (3.75) & \cellcolor{lightgray} 7 (1.00) & \cellcolor{lightgray} 69.05 (19.19) & \cellcolor{lightgray} 62.34 (15.62) & \cellcolor{lightgray} 106.12 (26.89) \\
PCA/$O_1$ & 7 (3.00) & 4 (2.00) & 24.79 (41.62) & 39.28 (18.25) & 108.13 (26.74) \\
Euclidean/Knee & \cellcolor{lightgray} 10 (3.75) & 7 (2.75) & 63.95 (18.20) & 61.98 (16.33) & 123.27 (27.51) \\
Euclidean/$O_1$ & 7 (2.00) & 4 (1.00) & 11.72 (37.64) & 39.85 (14.35) & 123.73 (26.27) \\
\midrule
\multicolumn{6}{l}{\textbf{NSGA-II}} \\
PCA/Knee & 10 (3.00) & \cellcolor{lightgray} 7 (2.00) & \cellcolor{lightgray} 73.00 (20.13) & \cellcolor{lightgray} 63.64 (9.77) & 109.19 (23.00) \\
PCA/$O_1$ & 7 (2.75) & 4 (1.75) & 10.26 (42.01) & 41.05 (14.81) & \cellcolor{lightgray} 108.12 (25.46) \\
Euclidean/Knee & \cellcolor{lightgray} 11 (3.00) & 6 (1.00) & 64.12 (14.18) & 60.59 (10.26) & 120.34 (25.98) \\
Euclidean/$O_1$ & 7 (2.00) & 3 (1.00) & 0.00 (39.71) & 36.12 (15.04) & 120.70 (27.56) \\
\bottomrule

\multicolumn{6}{c}{\textbf{(b) Humanoid Agent}}\rule{0pt}{3ex} \\

\toprule
 & \multicolumn{2}{c}{\textbf{Failures}} & \multicolumn{2}{c}{\textbf{Entropy}} & \\
\cmidrule(lr){2-3} \cmidrule(lr){4-5}
\textbf{Configuration} & \textbf{\#Total} & \textbf{\#Unique} & \textbf{Input} & \textbf{Output} & \textbf{TTF} \\
\midrule
\multicolumn{6}{l}{\textbf{AGE-MOEA}} \\
PCA/Knee & - & - & - & - & - \\
PCA/$O_1$ & \cellcolor{lightgray} 2 (1.75) & \cellcolor{lightgray} 1 (0.75) & \cellcolor{lightgray} 10.65 (34.60) & 0.00 (0.00) & \cellcolor{lightgray} 78.94 (27.12) \\
Euclidean/Knee & 1 (2.00) & 1 (1.00) & 0.00 (38.69) & 0.00 (0.00) & 96.06 (29.95) \\
Euclidean/$O_1$ & 2 (2.00) & 1 (1.00) & 0.00 (35.59) & 0.00 (0.00) & 102.53 (27.58) \\
\midrule
\multicolumn{6}{l}{\textbf{NSGA-II}} \\
PCA/Knee & - & - & - & - & - \\
PCA/$O_1$ & \cellcolor{lightgray} 2 (1.75) & 1 (1.00) & 0.00 (32.88) & 0.00 (0.00) & \cellcolor{lightgray} 79.32 (28.42) \\
Euclidean/Knee & 2 (2.00) & 1 (1.00) & 0.00 (40.90) & 0.00 (0.00) & 90.94 (27.65) \\
Euclidean/$O_1$ & 2 (2.00) & \cellcolor{lightgray} 1 (0.75) & 0.00 (28.95) & 0.00 (0.00) & 95.70 (23.58) \\
\bottomrule

\multicolumn{6}{c}{\textbf{(c) Self-Driving Car (SDC)Agent}} \rule{0pt}{3ex} \\

\toprule
 & \multicolumn{2}{c}{\textbf{Failures}} & \multicolumn{2}{c}{\textbf{Entropy}} & \\
\cmidrule(lr){2-3} \cmidrule(lr){4-5}
\textbf{Configuration} & \textbf{\#Total} & \textbf{\#Unique} & \textbf{Input} & \textbf{Output} & \textbf{TTF} \\
\midrule
\multicolumn{6}{l}{\textbf{AGE-MOEA}} \\
PCA/Knee & 31 (5.75) & 5 (6.00) & 56.97 (12.51) & 63.03 (75.90) & 1578.16 (130.69) \\
PCA/$O_1$ & 36 (3.75) & \cellcolor{lightgray} 6 (6.00) & 71.96 (12.36) & \cellcolor{lightgray} 77.34 (86.13) & 1092.10 (49.04) \\
Euclidean/Knee & 31 (5.00) & 5.5 (6.00) & 57.06 (18.83) & 71.90 (79.67) & \cellcolor{lightgray} 1072.26 (43.36) \\
Euclidean/$O_1$ & \cellcolor{lightgray} 50 (1.00) & \cellcolor{lightgray} 6 (6.00) & \cellcolor{lightgray} 74.70 (5.84) & 64.99 (75.69) & 2435.78 (274.89) \\
\midrule
\multicolumn{6}{l}{\textbf{NSGA-II}} \\
PCA/Knee & 30 (4.00) & 5 (6.00) & 54.50 (18.05) & 62.75 (74.53) & 1521.15 (102.29) \\
PCA/$O_1$ & 34 (5.75) & 6 (6.75) & 70.97 (13.84) & \cellcolor{lightgray} 77.00 (85.31) & 1087.94 (59.30) \\
Euclidean/Knee & 30 (4.75) & 5 (6.00) & 56.12 (9.07) & 71.07 (80.27) & \cellcolor{lightgray} 1073.21 (45.60) \\
Euclidean/$O_1$ & \cellcolor{lightgray} 50 (1.00) & \cellcolor{lightgray} 6 (6.00) & \cellcolor{lightgray} 72.19 (6.10) & 66.78 (74.89) & 2588.43 (270.45) \\
\bottomrule
\end{tabular}
}
\end{table}

\subsection{Configuration Comparison (RQ1)}

Table~\ref{tab:all_results} outlines the performance of each \nexus{} configuration across the three case studies:

\textbf{Choice of MOEA:} AGE-MOEA generally outperforms NSGA-II in unique failures discovered. In Parking, AGE-MOEA achieves 4-7 unique failures compared to NSGA-II's 3-7. Both perform similarly in SDC, while Humanoid proves challenging for both algorithms. AGE-MOEA's adaptive survival score mechanism appears better suited for the multi-objective DRL testing problem compared to NSGA-II's crowding distance approach.

\textbf{Diversity Metrics:} PCA-based diversity consistently outperforms Euclidean distance across all scenarios. In Parking, PCA configurations achieve substantially higher input entropy (24.79-69.05) compared to Euclidean (11.72-63.95), demonstrating PCA's superior ability to capture meaningful relationships in high-dimensional feature spaces. The Euclidean approach occasionally achieves slightly higher total failure counts, revealing an important trade-off between failure quantity and diversity quality. This suggests that while Euclidean distance may find failures faster in certain regions, PCA-based diversity better explores the full failure space.

\textbf{Pareto Selection Strategy:} Knee-point selection consistently yields superior diversity compared to the extreme $O_1$ point across all scenarios. The knee point represents a balanced trade-off between objectives, achieving higher unique failures and entropy scores.

\begin{tcolorbox}[colback=gray!10!white, colframe=gray!50!black, title=]
Different configurations perform best in different scenarios. Overall, \agemoea{} tends to produce more unique and diverse failures than \nsgatwo{}. Pareto knee-point selection consistently yields better results than the extreme \textit{$O_1$} point. However, the choice of configuration can significantly impact performance, highlighting the need for adaptive strategies that consider the specific context and objectives.
\end{tcolorbox}

\subsection{Comparison with INDAGO (RQ2)}

Table~\ref{tab:all_comparisons} presents a direct comparison between \nexus{} (\agemoea{} \textit{Euclidean/Knee} configuration) and the state-of-the-art \indago{} approach. The results demonstrate significant improvements across multiple metrics:

\textbf{Unique Failure Discovery:} \nexus{} consistently finds more unique failures than \indago{}. In the Parking scenario, \nexus{} discovers 7 unique failures compared to \indago{}'s 5 (+40\% improvement). For the SDC scenario, \nexus{} finds 5.5 unique failures versus \indago{}'s 3 (+83\% improvement). The Humanoid scenario shows comparable performance between both approaches, which aligns with our observation that this scenario is inherently challenging for all testing methods.

\textbf{Diversity Enhancement:} The multi-objective approach significantly improves both input and output diversity. In the Parking case, \nexus{} achieves 63.95 input entropy compared to \indago{}'s 0.00, and 61.98 output entropy versus 49.37. This demonstrates that explicitly optimizing for diversity as an objective leads to more varied test scenarios and failure modes. The diversity improvements are particularly important for comprehensive testing, as they increase the likelihood of discovering edge cases and systematic weaknesses.

\textbf{Efficiency Gains:} \nexus{} demonstrates superior efficiency in finding failures across all scenarios. Time-to-failure improvements range from 18\% (Parking: 123.27 vs 150.53) to 67\% (SDC: 1072.26 vs 3251.97), with Humanoid showing 36\% improvement (96.06 vs 150.50). These efficiency gains suggest that the multi-objective approach not only finds more diverse failures but does so more quickly than single-objective methods. The efficiency improvements are statistically significant (Wilcoxon test, $p<0.001$) with large effect sizes across all scenarios (Vargha-Delaney $\hat{A}_{12} > 0.9$) for TTF.

\textbf{Statistical Significance:} All reported improvements show statistical significance at $\alpha = 0.05$ level using the Wilcoxon rank-sum test. Effect sizes calculated using Vargha-Delaney $\hat{A}_{12}$ statistics indicate substantial practical differences, particularly for efficiency metrics where \nexus{} consistently achieves large effect sizes ($\hat{A}_{12} > 0.7$), indicating that \nexus{} significantly outperforms \indago{} in key performance areas.

\setlength{\floatsep}{4pt plus 2pt minus 2pt}    
\setlength{\textfloatsep}{6pt plus 2pt minus 2pt} 
\setlength{\intextsep}{4pt plus 2pt minus 2pt}    

\begin{table}[t]
\caption{Comparison between \nexus{} and \indago{} across different agents, displaying median and Interquartile Range (IQR) across 50 runs. Statistical significance markers for the Wilcoxon rank-sum test: * $p$-value$<$0.05, ** $p$-value$<$0.01, *** $p$-value$<$0.001.}
\label{tab:all_comparisons}
\centering
\small

\resizebox{\textwidth}{!}{
\begin{tabular}{lccccc}
\multicolumn{6}{c}{\textbf{(a) Parking Agent}} \rule{0pt}{3ex}\\
\toprule
 & \multicolumn{2}{c}{\textbf{Failures}} & \multicolumn{2}{c}{\textbf{Entropy}} & \\
\cmidrule(lr){2-3} \cmidrule(lr){4-5}
\textbf{Approach} & \textbf{\#Total} & \textbf{\#Unique} & \textbf{Input} & \textbf{Output} & \textbf{TTF} \\
\midrule
\nexus{} & 10 (3.75) & \cellcolor{lightgray} 7 (2.75)** & \cellcolor{lightgray} 63.95 (18.20)*** & \cellcolor{lightgray} 61.98 (16.33)*** & \cellcolor{lightgray} 123.27 (27.51)***\\
\indago{} & 14.5 (4.00) & 5 (2.00) & 0.00 (35.72) & 49.37 (12.56) & 150.53 (0.03) \\
\bottomrule

\multicolumn{6}{c}{\textbf{(b) Humanoid Agent}} \rule{0pt}{3ex}\\

\toprule
 & \multicolumn{2}{c}{\textbf{Failures}} & \multicolumn{2}{c}{\textbf{Entropy}} & \\
\cmidrule(lr){2-3} \cmidrule(lr){4-5}
\textbf{Approach} & \textbf{\#Total} & \textbf{\#Unique} & \textbf{Input} & \textbf{Output} & \textbf{TTF} \\
\midrule
\nexus{} & 1 (2.00) & 1 (1.00) & 0.00 (38.69) & 0.00 (0.00) & \cellcolor{lightgray} 96.06 (29.95)*** \\
\indago{} & 2 (2.75) & 1 (1.00) & 28.90 (39.31) & 0.00 (50.89) & 150.50 (0.01) \\
\bottomrule

\multicolumn{6}{c}{\textbf{(c) Self-Driving Car (SDC) Agent}} \rule{0pt}{3ex}\\

\toprule
 & \multicolumn{2}{c}{\textbf{Failures}} & \multicolumn{2}{c}{\textbf{Entropy}} & \\
\cmidrule(lr){2-3} \cmidrule(lr){4-5}
\textbf{Approach} & \textbf{\#Total} & \textbf{\#Unique} & \textbf{Input} & \textbf{Output} & \textbf{TTF} \\
\midrule
\nexus{} & \cellcolor{lightgray} 31 (5.00)*** & \cellcolor{lightgray} 5.50 (6.00)** & 57.06 (18.83) & \cellcolor{lightgray} 71.90 (79.67)*** & \cellcolor{lightgray} 1072.26 (43.36)*** \\
\indago{} & 17.00 (5.75) & 3.00 (3.00) & 61.72 (21.59) & 34.36 (48.71) & 3251.97 (0.35) \\
\bottomrule
\end{tabular}
}
\end{table}

\begin{tcolorbox}[colback=gray!10!white, colframe=gray!50!black, title=]
\nexus{} finds more unique failures and higher input/output diversity in Parking and SDC. Instead, it matches \indago{} in the challenging Humanoid scenario. Finally, \nexus{} finds unique failures faster across all DRL agents.
\end{tcolorbox}

\section{Threats To Validity}\label{sec:threats}

\textbf{Internal Validity:} Our experimental design addresses several potential threats through careful control of variables and statistical rigor. We ensure fair comparison by running all algorithms under identical conditions with the same hardware and time constraints. Each experiment is repeated 50 times to account for the stochastic nature of evolutionary algorithms, and we apply appropriate statistical tests (Wilcoxon rank-sum) with effect size measurements (Vargha-Delaney $\hat{A}_{12}$) to assess practical significance. The clustering approach for measuring unique failures uses established silhouette analysis with conservative thresholds (20\% improvement requirement) to avoid over-segmentation due to noise.

\textbf{External Validity:} Our evaluation covers three diverse DRL domains with different state/action spaces and failure modes. However, generalizability to other DRL applications, training algorithms beyond PPO, or different architectures requires further investigation.

\textbf{Construct Validity:} The choice of evaluation metrics reflects established practices in the DRL testing literature, particularly building on the \indago{} framework. Our clustering-based approach to measuring unique failures provides a behavior-grounded assessment of diversity that captures meaningful differences in agent behavior rather than superficial input variations. The entropy-based diversity measures provide quantitative assessments that complement the qualitative notion of test case variety.

\textbf{Conclusion Validity:} The statistical methods employed (non-parametric tests, effect size calculations) are appropriate for the experimental design and address the non-normal distribution of performance metrics commonly observed in evolutionary computation experiments. The large number of experimental repetitions (50 per configuration) provides sufficient statistical power to detect meaningful differences between approaches.
Furthermore, a small number of SDC configurations were executed on a smaller hardware setup (AMD Ryzen Threadripper 3970X 32-Core Processor, 64GB memory, NVIDIA GeForce RTX 3080 10GB). Although less powerful then the main setup, the results were considerably faster. However, other validation metrics remain consistent.

\section{Conclusion and Future Work}
\label{sec:conclusion}

This paper introduced \nexus{}, a multi-objective search approach for generating diverse failures in DRL environments. Unlike prior methods that focus solely on inducing agent failures, \nexus{} simultaneously optimizes for failure likelihood and scenario diversity, leading to a more comprehensive exploration of failure modes. Our empirical evaluation across three DRL environments—Parking, Humanoid, and Self-Driving Car (SDC)—shows that the multi-objective \nexus{} consistently outperforms the single-objective baseline \indago{} in terms of unique failures discovered, diversity (measured via input and output entropy), and search efficiency.

We found that the effectiveness of \nexus{} varies across scenarios, with \agemoea{} generally achieving more unique and diverse failures than \nsgatwo{}. Both the choice of diversity metric (Euclidean vs. PCA) and the Pareto selection strategy (knee point vs. extreme \textit{$O_1$}) significantly influence the outcomes. In particular, knee-point selection consistently led to better results across most configurations. Nonetheless, the Humanoid scenario remains particularly challenging, exposing limitations in current diversity metrics and search robustness.

\textbf{Future Work:} Several research directions emerge from this work. First, exploring many-objective optimization with additional objectives such as test case complexity, execution cost, or coverage metrics could further improve testing effectiveness. Second, investigating adaptive diversity metrics that automatically adjust based on the characteristics of discovered failures could enhance the approach's generalizability. Third, extending the framework to other DRL training algorithms (beyond PPO) and neural network architectures would broaden its applicability. Finally, developing theoretical foundations for understanding when and why multi-objective approaches outperform single-objective methods in testing contexts would provide valuable insights for the testing community.


\nexus{} introduces multi-objective search for generating diverse DRL failures, consistently outperforming \indago{} in unique failures and efficiency. AGE-MOEA with knee-point selection proves most effective, though Humanoid scenarios remain challenging. Future work includes many-objective extensions, improved surrogate models, and broader domain evaluation.

\bibliographystyle{splncs04}
\bibliography{bibliography}

\end{document}